\documentclass[preprint,review,5p,times,twocolumn]{elsarticle}

\usepackage{lineno,hyperref}
\modulolinenumbers[20]
\usepackage{caption,subcaption}
\usepackage{multirow}
\usepackage{bm} 
\usepackage{graphicx}
\usepackage{amsmath}
\usepackage{caption}
\usepackage{changepage}

\biboptions{sort&compress}
\begin{document}

\begin{frontmatter}

\title{6-DOF Feature based LIDAR SLAM using ORB Features from Rasterized Images of 3D LIDAR Point Cloud}

\author{Waqas Ali} 
\author{Peilin Liu} 
\author{Rendong Ying} 
\author{Zheng Gong}

\address{309 Microelectronics Building, Shanghai Jiaotong University, Shanghai} 

\begin{abstract}
	An accurate and computationally efficient SLAM algorithm is vital for modern autonomous vehicles. To make a lightweight algorithm,
	most SLAM systems rely on feature detection from images for vision SLAM or point cloud for laser-based methods. Feature detection through a 3D point cloud becomes a computationally challenging task. In this paper, we propose a feature detection method 	by projecting a 3D point cloud to form an image and apply the vision-based feature detection technique. The proposed method gives repeatable and stable features in a variety of environments. Based on such features, we build a 6-DOF SLAM system consisting 	of tracking, mapping and loop closure threads. For loop detection, we employ a 2-steps approach i.e. nearest key-frames detection and loop candidate verification by matching features extracted from rasterized LIDAR images. Furthermore, we utilize a key-frame structure to achieve a lightweight SLAM system. The proposed system is evaluated with implementation on the KITTI dataset and the University of Michigan Ford Campus dataset. Through experimental results, we show that the algorithm presented in this paper can substantially reduce the computational cost of feature detection from the point cloud and the whole SLAM system while giving accurate results.
\end{abstract}


\end{frontmatter}

\linenumbers

\section{Introduction}

Autonomous mobile robots are becoming an integral part of
our lives due to their ability to navigate through complex and
dangerous environments with increased efficiency. To enable
a vehicle to move autonomously it must have a map of the environment and positioning ability within the map. The map of
the surrounding through which a vehicle needs to traverse is critical for autonomous operation for several reasons. First, it can
be used for path planning of the robot. Second, it can be used
to reduce localization error. When a robot revisits an area, if
the map is saved it can be used to reset the position of the robot and
correct the whole trajectory. This process is called loop closure.
For some applications such as inside a house or small factory,
it might be possible to save a pre-built map. But it becomes a
hectic task as the path length grows. Therefore, our system must build a map as it moves through an area.

SLAM \cite{Ref1} \cite{Ref2} problem is associated with the localization of a moving body and building an accurate map of the environment.
An accurate and computationally efficient algorithm for simultaneous localization and mapping is an integral goal for modern
autonomous vehicles. SLAM is commonly implemented using
2 types of sensors i.e. vision or laser. LIDAR sensors have several advantages over vision-based sensors; it provides highly
accurate range data and it is not affected by light conditions.
By using LIDAR, it is possible to achieve higher accuracy for
localization and mapping.

A modern 3D LIDAR sensor is composed of rotating laser beams to acquire 3D range data of the environment. Simultaneous localization and mapping with laser sensor is achieved mostly in 2 ways. The first method is the use of scan matching using the whole point cloud to estimate the pose of a robot. This method presents some advantages i.e. better accuracy and it is applicable in any environment, but matching a dense point cloud becomes computationally expensive.

The other method is based on scan registration using features extracted directly from point clouds. Feature-based methods present better computational efficiency. But most of the feature-based methods employed in literature are environment-specific. For instance, corner and plane based feature detectors were designed for indoor environments \cite{Ref5, Ref6, Ref7} and tree detectors have been applied in outdoor settings \cite{Ref8,Ref9}. These feature detectors are designed for a specific environment and there has been little work for general-purpose feature extraction from 3D laser point cloud for SLAM application. 

Our system builds on the approach of Li and Olsen \cite{Ref22} i.e. general-purpose feature detection using rasterized LIDAR images. Our main contributions are as follow:

\begin{enumerate}[I.] 
	\item We present a novel feature detection algorithm for the 3D LIDAR point cloud. By the application of the ORB feature detection method on rasterized LIDAR images, we increase the computational efficiency of feature detection while giving repeatable and stable features. Instead of searching through the whole point cloud for feature points, we only need to search through the rasterized image.
	\item Using ORB features from rasterized LIDAR images, we can estimate the 6DOF pose and build an accurate map of the environment. 
	\item A novel loop closure detection technique for the LIDAR based SLAM system is introduced in this paper. We search for the nearest key-frames and then match features to detect loop closure.
\end{enumerate}

In our system, accurate results are provided by multiple levels of optimization. Firstly, the vehicle's pose is estimated and optimized using the least square method. Also, for low feature scenarios pose graph optimization is performed with visual odometry constraints. Next, both trajectory and local maps are optimized using local bundle adjustment, and lastly employing loop closure detection, the whole trajectory is optimized. We use the same features, used for tracking and mapping threads, to detect loop closure which makes our approach simple and efficient. In this paper with the help of extensive experiments, we show the efficiency of the proposed algorithm. Using the KITTI  dataset \cite{Ref32} and the University of Michigan Ford Campus dataset \cite{Ref33}, our method gave results on par with state of the art algorithms at a much lower computational cost.

The rest of the paper is organized as below, Section-2 gives a brief survey of LIDAR based SLAM algorithms divided into three sections i.e. ICP based methods, feature-based methods and loop closure detection. Then we give a brief overview of our system in section-3. Section-4 to section-9 describe in detail the working of each thread of our algorithm i.e. Pre-processing, Rasterization, Visual Odometry, Tracking, Mapping and Loop closure. Section-10 deals with the experimental implementation of our algorithm with KITTI \cite{Ref32} and the University of Michigan Ford Campus dataset \cite{Ref33}. The results are analyzed and compared with state-of-the-art methods.

\section{Related Work}

\subsection{ICP methods}

For most laser SLAM algorithms, scan matching is used to estimate the relative motion between two scans and ICP algorithms \cite{besl1992method,chen1992object} are commonly used for this purpose. It is a simple technique and easy to implement. ICP algorithm starts with an initial transformation to find correspondences between the 2 point clouds and iterate until the solution converges. The basic ICP algorithm requires a good initial estimate to ensure accurate results. To make ICP more efficient several variants have been proposed in the literature \cite{censi2008icp,magnusson2007scan,lu1997robot} . Pomerleau et al. \cite{Ref10} proposed a comparison framework between the two most popular ICP algorithms for point cloud registration i.e. point-to-point and point-to-plane. ICP based methods can produce accurate results depending on a good initial guess. But these algorithms are computationally expensive.

\subsection{Feature detection}

To achieve better computational efficiency, most modern laser SLAM methods extract features from the point cloud to perform scan registration. Dong and Barfoot \cite{Ref12} proposed a method of extracting visual features from intensity and range images acquired from AUTONOSYS LIDAR. Other similar approaches \cite{Ref13,Ref14} estimate the motion by matching features from images and computing motion model from Gaussian processes. These methods also employ AUTONOSYS LIDAR for experimental evaluation. This LIDAR combines laser and camera to form intensity and range images and requires a dense point cloud. Whereas we propose a method that only requires a 3D point cloud to form rasterized images and doesn't need a dense point cloud. Steder et al. \cite{Ref21} used point features extracted from panoramic range images for place recognition. To form panoramic images from point clouds, dense scans are required that must be collected in a stop-scan-go manner. It makes this feature detection technique insufficient for SLAM applications. Zhuang \cite{Ref20} introduced a method of transforming point cloud data to a bearing angle image and then extract SURF based features. It also requires a dense point cloud to form such images, which is not a necessity for our proposed method.

Chong et al. \cite{chong2013mapping} proposed a method to form a synthetic 2D LIDAR for the localization of a robot in a 3D environment. Vertical planes are extracted from the 3D point cloud and projected to a virtual plane to form synthetic 2D LIDAR and then Monte-Carlo localization is applied to this synthetic 2D LIDAR data. In the process of projecting 3D points to 2D planes using such a technique, a lot of information is lost which affects the final result. Zhang and Singh \cite{Ref19} extracted corners and plane features directly from the point cloud and were able to estimate accurate odometry in real-time. For such a technique, the whole point cloud needs to be scanned for feature detection. While we only need to scan an image to extract features.   

Our approach is most similar to the work of Li and Olsen \cite{Ref22}. In their work, they introduced the idea of extracting general-purpose feature detectors by applying image processing methods on a rasterized image of the point cloud. Their method used a multi-scale Kanade – Tomasi corner detector to extract features, but their work was only limited to feature extraction. We take this idea further and use a pinhole camera model for image formation instead of direct rasterization. This permits us to apply advanced feature detectors such as ORB and we can detect stable and repeatable features from such images. During the process of image formation, the z-value of each 3D point is saved in the form of pixel intensity. This allows us to estimate accurate 3D pose and build a precise map.

\subsection{Loop closure detection}

For robust SLAM applications, loop closure detection is vital, as it corrects the accumulated error in the estimated trajectory over time. Hess et al. \cite{Ref24} present the method of dividing the map into sub-maps and use branch and bound approach for computing scan to sub-map matches as constraints in the pose graph structure. The optimization is done quickly to ensure operation in real-time. Dub{\'e} et al. \cite{Ref25} propose a point cloud segmentation-based approach. An incoming point cloud is divided into segments, its descriptors are extracted and saved into the system. The new segments are matched and after geometric verification loop closure candidate is selected. Magnusson et al. \cite{Ref26} propose an approach where global statistics are computed for point cloud and a simple threshold-based approach is used to compute loop closure candidates. Behley and Stachniss \cite{Ref27} propose a surfel-based mapping approach for building global maps and use a map-based criterion for loop closure detection.

Using ORB features in our approach gives us an edge over existing methods i.e. loop closure candidate verification based on feature matching is simple and more efficient. In our algorithm loop closure detection is performed in two steps. First, a loop closure candidate is selected based on the nearest key-frame detection. Then further verification is performed based on feature matching. If these two steps are passed the loop closure constraint is added to the pose-graph structure and the trajectory is optimized.

\begin{figure*}[ht]
	\centering
	\includegraphics[scale=0.5]{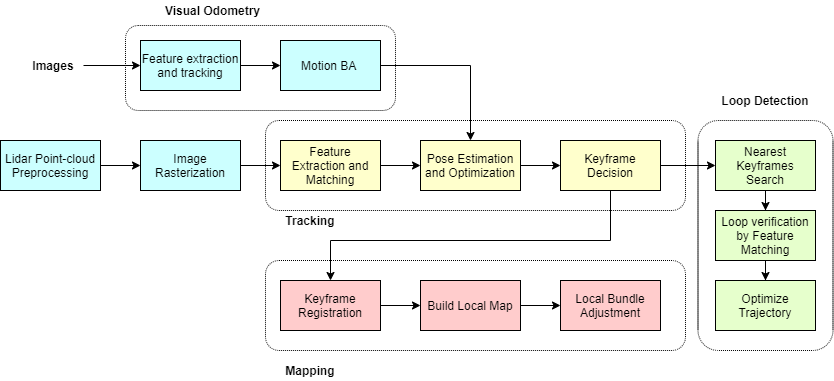}
	\caption{System overview, showing LIDAR point to image formation, and steps involved in tracking, mapping and loop closure threads.} 
	\label{fig:system}
\end{figure*}

\section{System Overview}

The algorithm starts with the pre-processing of LIDAR data. We use RANSAC based plane fitting to detect and remove the ground plane from the point cloud. The remaining point cloud is then given to the image rasterization node. The projection of the point cloud to form an image is an integral part of our feature detection approach. Figure~\ref{fig:system} shows an overview of the algorithm. We used the pinhole camera model to project every point to the image plane and in this way, a rasterized image is formed. The depth information is preserved in the process of rasterization by assigning the intensity of each point on a rasterized image equal to the z-value of that point in world coordinates. After the point cloud has been rasterized to form an image, the rest of the algorithm is divided into 3 main thread: Tracking, Mapping, and Loop Closure.

Visual odometry thread works in parallel with the tracking. Image features are detected and tracked, then motion BA is performed to give pose in the local frame. The tracking thread takes rasterized images and extracts ORB features from these images. Features between scans are matched and outliers are removed using RANSAC. The matched feature points are projected back to LIDAR coordinates. We estimate ego-motion using an efficient ICP method based on these feature points with known correspondences. Based on the motion transformation, the pose of LIDAR is calculated. A feature consistency check is performed and if very low features are detected, pose graph optimization is performed with visual odometry constraints. To make our system more efficient, we use a key-frame structure. The tracking thread is also responsible to decide when to add a new key-frame. The details of the strategy for the new key-frame decision are discussed in the tracking section. Once a new key-frame has been detected it is passed on to the key-frame database. 

The mapping thread registers new key-frames and at the same time using pose information tracks and saves a local map. In this thread, we build a pose graph structure with current $n$ key-frames and map-points of the local map and apply bundle adjustment to optimize key-frame poses and map-points. To keep the bundle adjustment efficient, we employ a culling algorithm to remove redundant key-frames and map-points. 

The third thread deals with loop closure detection and correction of the trajectory. The loop closure thread saves each key-frame information from tracking and whenever a new key-frame is added, it searches for the nearest key-frames and then uses feature matching to finds loop closure. When loop closure is detected, a pose graph is formed with all key-frames as nodes, and loop closure constraint is added to the graph and optimized to correct the drift in trajectory.

\section{Pre-processing}

It is important to remove outliers from the point cloud so that after it is rasterized to form an image, we can extract stable features. For a point cloud recorded by a 3D LIDAR, circular rings are formed on the ground, as a result of the rotating motion of multi-line LIDAR. If a rasterized image is formed with these rings, we are unable to extract stable features and the motion estimation is distorted. We apply Random Sample and Consensus (RANSAC) \cite{Ref28} plane fitting to detect and remove the ground plane and the rest of the point cloud is then rasterized to form an image. 

\begin{figure}[htp!]
	\centering
	\includegraphics[scale=0.5]{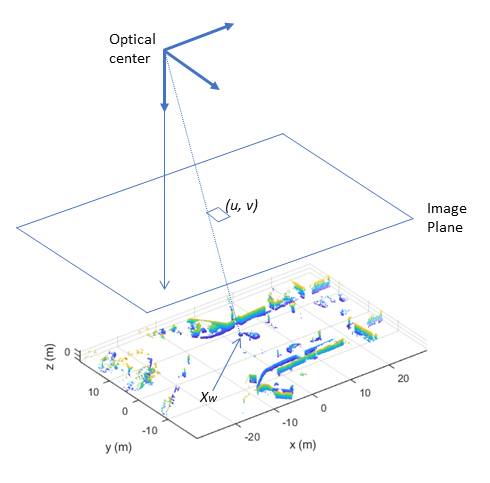}\\
	\caption{The process of projecting a 3D point from the LIDAR point cloud on to the image plane. Here z-axis is set as the optical axis. }
	\label{fig:camera}
\end{figure}

\section{Rasterization \vspace{0em}}

In our system, we use a pinhole camera model to project the point cloud to an image plane. Figure~\ref{fig:camera} shows the projection of 3D points to the image plane. The goal of this process is to map a 3D point to get a 2D pixel point with intensity value. The image formed as a result of using the pinhole camera model instead of direct rasterization is suitable for the application of advanced feature detection methods such as ORB.

The process of image formation is simple; the point cloud is given in the LIDAR coordinates, it is first transformed to camera coordinates and then projected to the image plane. For a point in LIDAR coordinates $P_l$ can be transformed to $P_c$ in camera coordinates using the following equation:

\begin{align}
P_c = \bm{R} (\bm{I} | \bm{t})P_l
\label{eq.1}
\end{align}

Equation\ref{eq.1} shows the extrinsic camera parameter, it is given as a rotation matrix $\bm{R}$ and a translation vector $\bm{t}$. Then we should define the camera intrinsic parameters, which map a 3D point in the camera coordinates to the image plane. These parameters are given as:
\begin{align}
\left[
\begin{array}{c}
u \\
v \\
w \\
\end{array}
\right]
= 
\left[
\begin{array}{ccc}
f & 0 & t_u \\
0 & f & t_v \\
0 & 0 & 1 \\
\end{array}
\right]
\left[
\begin{array}{c}
X \\
Y \\
Z \\
\end{array}
\right]
\label{eq.2}
\end{align}

\noindent
Here $f$ is the focal length and $(t_u, t_v)$ is the optical center of the camera. Point $P_c = [X \quad Y \quad Z]^T$ is a point in camera coordinates and its projection on the image plane is $P_i = [u \quad v \quad w]^T$. Combining equation ~\ref{eq.1} and ~\ref{eq.2}, we get a full camera projection model
\begin{align}
P_i = \bm{K}\bm{R} (\bm{I} | \bm{t}) P_l = \bm{C}P_l
\label{eq.3}
\end{align}

\noindent
The matrix $\bm{K}$ defines the intrinsic parameters of a camera.

\begin{figure}[h]
	
	\begin{subfigure}{.25\textwidth}
		\centering
		\includegraphics[scale=0.4]{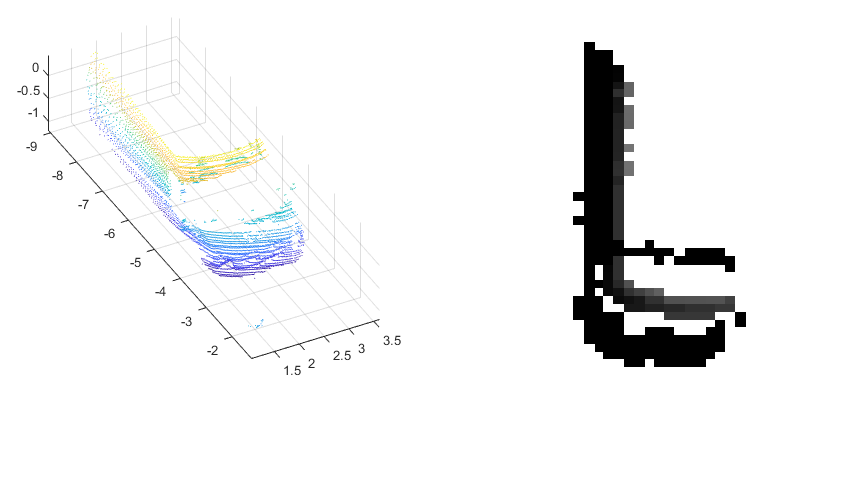}
		\label{fig:sub1}
	\end{subfigure}%
	\hfill
	\begin{subfigure}{.25\textwidth}
		\centering
		\includegraphics[scale=0.4]{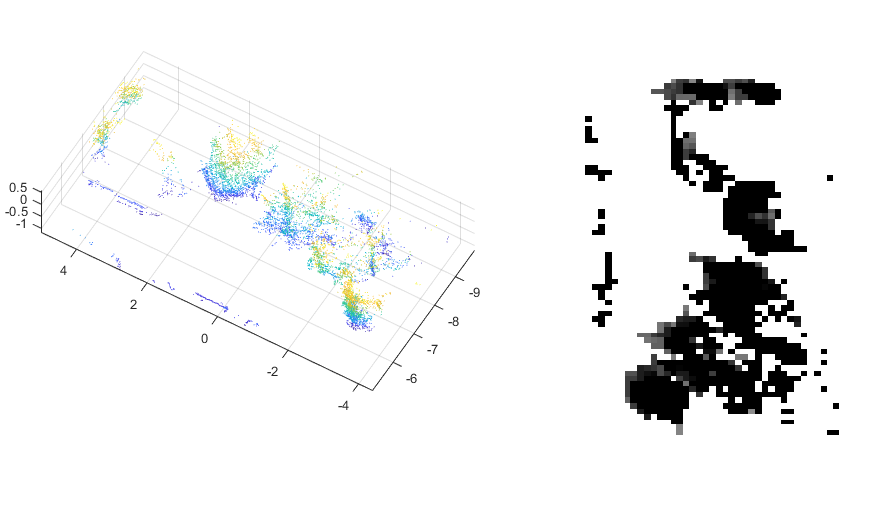}
		\label{fig:sub2}
	\end{subfigure}
	\caption{Two examples of point cloud rasterization are shown here. Point clouds from a tree and car include points with different heights and these values correspond to the intensity of pixels in the image.}
	\label{fig:raster_im}
\end{figure} 

Using equation~\ref{eq.3} we project the point cloud to an image plane to form a gray-scale image. An important step is to save the $z-axis$ value of each 3D point that gives the rasterized image some physical meaning, minimize loss of information and minimize the re-projection error of feature points. For that purpose, we assign the intensity value of each 2D point on the image plane equal to the z-axis value of its corresponding 3D point. An example of a laser point cloud projection to form an image is shown in figure~\ref{fig:raster_im}, which shows the gray-scale images for a car and trees. Another issue is that for a laser scan there might be several points existing at the same $(X,Y)$ location with different z-values. So during the projection of such points to the image plane, with an optical axis set to the z-axis, they form a single pixel. In such cases, we assign the maximum z-value of such points to the corresponding pixel on the rasterized image. The reason for using the maximum $Z$ value is that it represents the visible height for such points by LIDAR.

\begin{figure}[h]
	\centering
	\includegraphics[scale=0.5]{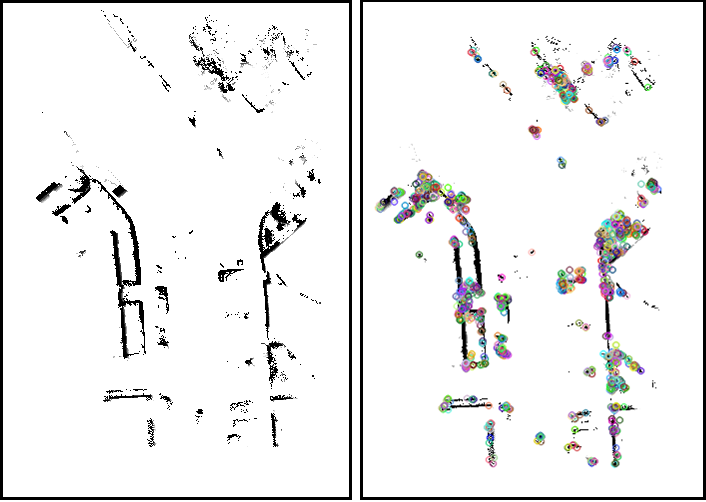}\\
	\caption{On the left is an image obtained after rasterization along the z-axis and on the right are ORB features detected from this image}
	\label{fig:features}
\end{figure} 

\section{Visual Odometry}

There are several states of the art visual odometry methods present in the literature. We design the visual odometry thread based on the method presented in \cite{orb}. As this part is not one of the main contributions of this paper, we provide a brief description of the VO method. VO is performed in two steps i.e. first features extraction and tracking to estimate camera pose and then perform local BA to optimize the pose. For each new image, FAST corners are detected, if enough features are not found detector threshold is adapted. Next ORB descriptors are assigned for each corner \cite{Ref29}. The camera pose is calculated by matching these features. After tracking the pose, a search for map-points in the last frame is performed based on the motion model. After successfully finding correspondences, local BA is performed to optimize the pose. Frame by frame visual odometry poses are fed to the tracking thread, where it is used to optimize the lidar pose. The main purpose of VO is to ensure that the tracking thread does not fail in low feature environments.

\section{Tracking}

\subsection{Feature Extraction}

\begin{figure}[h]
	\centering
	\includegraphics[scale=0.5]{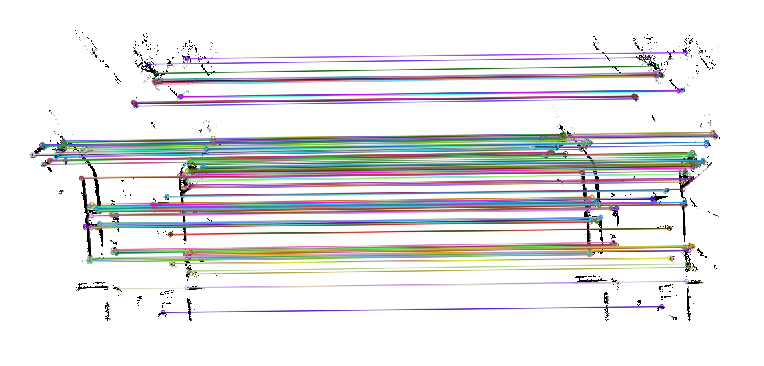}\\
	\caption{For a frame to frame motion estimation, features are matched from both frames and after removing outliers these features are re-projected to LIDAR coordinates for motion estimation.}
	\label{fig:matching} 
\end{figure}

A Rasterized image is given to the tracking thread, where the first step is feature extraction. First Gaussian blur is applied to smooth the rasterized image and then ORB features \cite{Ref29} are extracted from this image.  ORB consists of a FAST corner detector and oriented BRIEF descriptor. We apply the FAST corner detector on the rasterized images and then oriented BRIEF descriptors are computed for each corner. Figure~\ref{fig:features} shows ORB features detected from a rasterized LIDAR image. Next, we find correspondences between consecutive scans by matching features and remove outliers using the RANSAC algorithm \cite{Ref28}. Figure~\ref{fig:matching} shows matched features from rasterized images obtained from 3D LIDAR data. As mentioned in the last section z-values of each 3D point are saved during image formation. Using this information and camera parameters, a matched set of features is projected back to LIDAR coordinates. These points are then used for pose estimation.  

\subsection{Pose Estimation and Optimization}

The next step is to estimate the motion from the feature points and optimize the pose. Assuming, we have a set of feature points extracted from rasterized images projected back to LIDAR coordinates and given as $p_t$ and $p_{t-1}$. Our objective function to estimate motion is defined as follow: 

\begin{equation}
f\left(\bm{Rot},\bm{tr}\right) = \frac{1}{N_{p_t}}\sum_{i=1}^{N_{p_t}} ||p_{t-1} - \bm{Rot}p_t - \bm{tr}||^2
\label{eq.4}
\end{equation}

ICP algorithm is used to estimate the rotation matrix $\bm{Rot}$ and translation vector $\bm{tr}$. ICP algorithm can give accurate results depending on a good initial guess to compute correspondences. But its main drawback is higher computational cost controlled by the number of points. In this paper, we can improve the efficiency of ICP by first using feature points instead of whole point-cloud and second providing correspondences between the points. After estimating the motion between 2 sets of points, the LIDAR pose is calculated. At this stage, a pose graph is built which consists of VO constraints. A features consistency check is performed and if very few features are matched, pose graph optimization is performed to update the robot pose with visual odometry in low feature scenarios. This vehicle pose is also used in mapping thread to track features in world coordinates.

\subsection{Key-frame Decision}

The last step of the tracking thread is to decide when a new key-frame should be selected. We keep our strategy for key-frame selection simple. The decision process only needs 2 requirements i.e.

\begin{enumerate}[A.]
	\item At least 5 frames have passed
	\item There are at most 100 points in common with the last key-frame
\end{enumerate}

Whenever a new key-frame $K_i$ is detected, the tracking thread passes this information to the key-frames database where all the key-frames are stored. These key-frames can be accessed by mapping and loop-closure threads. After the optimization is finished in both mapping and loop closure threads, key-frames pose values are updated with the optimized values. Each key-frame $K$ stores the following information:

\begin{enumerate}[A.]
	\item 6DOF pose
	\item 3D position of the feature points seen at that position
	\item Descriptors of these feature points
\end{enumerate}

\section{Mapping}

\subsection{Map-points Registration}

The mapping thread starts by building a pose graph containing key-frames and map-points as nodes. When a new key-frames $K_i$ is inserted into the database by tracking thread, its pose is added to the graph as a node. The feature points $P_i$ seen at that key-frame are first tracked using the key-frame pose and then used to build a local map. At the very first key-frame insertion $K_0$, the local map is initialized using all the points seen at key-frame $K_0$. When the next key-frame $K_1$ comes, its feature points are matched against the local map and the unmatched points are inserted into the local map. This process is carried out with every new key-frame. The key-frame poses and map-points in the local map are optimized using local bundle adjustment.

To limit the complexity of local BA, only a set of current key-frames are kept in the graph structure. This requires a culling strategy for key-frames and map-points in the local map. The culling scheme for key-frames in our algorithm is kept simple. During local bundle adjustment only current $n$ key-frames are optimized, so when a new key-frame is received the redundant key-frames are removed from the graph. At the same time, map-points culling is also done. After redundant key-frames are removed, a check for unnecessary map-points is performed. Any map-point in the local map that is not visible by any of the current key-frames is removed. The two culling techniques ensure that our algorithm remains lightweight and computationally efficient. 

\subsection{Bundle Adjustment}
Local bundle adjustment is used to optimize $n$ current key-frames and map-points inside the local map. The map-points are projected to LIDAR coordinates from world coordinates and then to camera coordinates. By solving the local BA problem, we want to minimize the re-projection error in the 3D positions of map-points. The key-frame pose $P_{iw}$ is LIDAR pose registered as a node in the graph. The transformation from LIDAR to camera coordinates remains fixed. For a point observed in an image $\bm{x}_{ij}$ and its corresponding map-points with 3D position $\bm{x}_{wj}$, the cost function to minimize re-projection error is given as:
\begin{equation}
e_{ij} = x_{ij} - \bm{C}(\bm{P}_{iw},\bm{x}_{wj})
\label{eq.5}
\end{equation}

The cost function is for a key-frame node $i$ connected through a constraint to a map-point node $j$. $\bm{C}$ represents a function to project a 3D point to an image as derived in equation~\ref{eq.3}. It includes the intrinsic and extrinsic camera parameters. Note that the camera projection function $\bm{C}$ defined in section-5 remains constant. Our goal is to minimize equation ~\ref{eq.5} to find the optimal configuration for node x.
\begin{equation}
F(x^*) = argmin \sum_{ij}e^T_{ij}\bm{\Omega}_{ij}e_{ij}
\label{eq.6}
\end{equation}

\noindent
Where $\bm{\Omega_{ij}}$ is the covariance matrix associated with the observation. We define the Jacobian $\bm{J}_{ij}$ to solve equation~\ref{eq.6} as follow

\begin{center}
	$\bm{J}_{ij} \quad = \quad \left(0...0 \quad \bm{A}_{ij}\quad 0...0 \quad \bm{B}_{ij} \quad 0...0\right)$		
\end{center}

\begin{center}
	$\bm{A}_{ij} \quad = \quad \frac{\partial e_{ij}}{\partial x_i} , \quad\quad
	\bm{B}_{ij} \quad = \quad \frac{\partial e_{ij}}{\partial x_j}$
\end{center}

\noindent
We find the derivative of the cost function in equation ~\ref{eq.5} with respect to node $i$ and $j$ to estimate the terms in Jacobian. For combining the camera projection parameters and LIDAR pose to get the relationship between a point in image and world coordinates, we get the following expression
\begin{equation}
\left[
\begin{array}{c}
wu \\
wv \\
w \\
\end{array}
\right]
= 
\left[
\begin{array}{ccc}
f & 0 & t_u \\
0 & f & t_v \\
0 & 0 & 1 \\
\end{array}
\right]
\left[
\bm{R} \quad \bm{t}
\right]
\left[
\bm{R}_{iw} \quad \bm{t}_{iw}
\right]
\left[
\begin{array}{c}
X_w \\
Y_w \\
Z_w \\
1 \\
\end{array}
\right]
\label{eq.7}
\end{equation}

\noindent
For our system $\bm{R}$ and $\bm{t}$ are camera extrinsic parameters, $\bm{R}$ is used as identity and $\bm{t}$ is constant. $\bm{R}_{iw}$ and $\bm{t}_{iw}$ are the rotation matrix and translation vector of LIDAR pose respectively. $f$ is the focal length of the camera and $(t_u, t_v)$ is the optical center of the camera. The left-hand side of the equation~\ref{eq.7} represents point $(u, v)$ on image written in homogeneous coordinates. In equation~\ref{eq.7} the terms to be optimized are $X_w, Y_w, Z_w$ and $\bm{R}_{iw}$ and $\bm{t}_{iw}$. The 2 terms of Jacobian are given as follow:

\begin{equation}
\bm{A}_{ij} = 
\left[
\begin{array}{c c c}
\frac{X}{Z} & 1 & 0 \\
\frac{Y}{Z} & 0 & 1 \\
\end{array}
\right]
\label{eq.8}
\end{equation}

\begin{equation}
\bm{B}_{ij} = 
\left[
\begin{array}{c c c}
\frac{f}{Z} & 0 & -\frac{fX}{Z^2} \\
0 & \frac{f}{Z} & -\frac{fY}{Z^2} \\
\end{array}
\right]
\label{eq.9}
\end{equation}  

\noindent
Once the local BA is complete, we update the key-frame poses and the optimized map-points are saved to the map.

\section{Loop-closure Detection}

\begin{figure}[h]
	\centering
	\begin{subfigure}{.45\textwidth}
		\centering
		\includegraphics[scale=0.25]{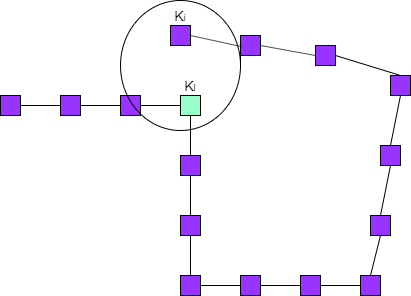}
		\caption{pose graph containing key-frames, $K_l$ falls inside the threshold distance of the current key-frame $K_i$}
		\label{fig:loop:sub1}
	\end{subfigure}%
	\hfill
	\begin{subfigure}{.45\textwidth}
		\centering
		\includegraphics[scale=0.25]{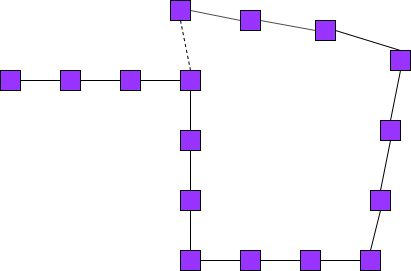}
		\caption{after verification with feature matching a constraint is added between $K_i$ and $K_l$}
		\label{fig:loop:sub2}
	\end{subfigure}
	\begin{subfigure}{.45\textwidth}
		\centering
		\includegraphics[scale=0.25]{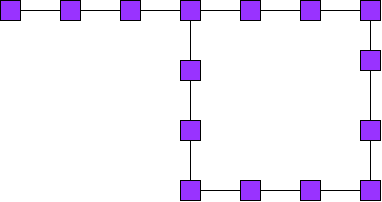}
		\caption{pose graph after optimization}
		\label{fig:loop:sub3}
	\end{subfigure}
	\caption{Illustration of pose graph optimization with loop closure constraints to correct the drift}
	\label{fig:loop}
\end{figure}

We use 2 steps approach to detect loop closure candidates i.e. candidate selection by searching for nearest key-frames and then use feature matching for candidate verification. Loop closure thread takes keyframe's information whenever a new key-frame is registered. Its pose is added as a node into the pose-graph. Contrary to local bundle adjustment, for loop closure optimization the pose-graph structure only contains key-frame poses as nodes. The constraints between these nodes are the estimated motion from the tracking thread. Due to the use of ORB features, we can detect loop closure more efficiently. Figure~\ref{fig:loop} shows the working of loop closure detection and pose-graph optimization with loop constraint. 

Although we are using a vision-based feature detection technique for rasterized images, these images and features in our system are quite different from the images of an actual camera. Keeping that in mind we devise an approach of loop closure detection suitable for feature points detected from LIDAR rasterized images. Whenever a new key-frame $K_i$ is added, we search for the nearest key-frames using a distance threshold. Figure \ref{fig:loop:sub1} shows pose graph with key-frames poses as nodes. A key-frame $K_l$ falls inside the distance threshold of key-frame $K_i$, then we match features seen at $K_i$ and $K_l$. First, we apply the ratio test \cite{Ref30} to check for outliers. If the first test is passed, then we use RANSAC to check for further outliers and verify the selected candidate. After final verification, a loop constraint is added between the respective nodes in the graph as shown in figure \ref{fig:loop:sub2}. We use the Levenberg Marquardt optimizer of GTSAM \cite{Ref31} library to optimize the graph. In figure \ref{fig:loop:sub3}  the resulting trajectory is shown after optimization is finished. Once optimization is finished all key-frames poses are updated.

\section{Experiments}

To evaluate the performance of our system we have implemented our algorithm on two public datasets i.e. KITTI dataset \cite{Ref32} and the University of Michigan Ford Campus dataset\cite{Ref33}. Our algorithm is written in C++ and tests are run on a laptop with an i5-8300H processor and 8GB RAM. To validate the performance of our system we use the open-source implementation of LOAM \cite{Ref19} and cartographer \cite{Ref24}. The performance is assessed in three steps i.e. tracking accuracy, loop closure performance and computational costs.

\subsection{Experimental Setup}
KITTI odometry benchmark dataset provides 22 sequences in total, comprising of image and LIDAR data. There are 11 training sequences provided with accurate ground truth and 11 test sequences. Data is gathered using a Velodyne HDL-64E laser sensor, four cameras, and an inertial and GPS navigation system installed on a car. The data provided is recorded from a variety of environments including urban, rural, and along the highway. In addition to the KITTI dataset, we also use the University of Michigan Ford Campus dataset. It uses HDL-64E LIDAR with Applanix GPS and an inertial system for ground truth. These two datasets are fitting to our requirement as it can show the performance of our feature detection method in different environments.

For experiments with the KITTI dataset, we use 11 training sequences from the KITTI odometry benchmark dataset to tune the parameters of our system. After testing the algorithm on training data, it is run on the remaining 11 test sequences and submitted to the KITTI odometry benchmark. In this paper, we compare the performance of our method with LOAM, which is the best performing laser-based method on the KITTI dataset. 

For the next step we use cartographer \cite{Ref24} to assess the loop closure detection performance of our system, it is a state-of-the-art laser SLAM algorithm with loop closure detection. Cartographer requires IMU data in addition to laser scans for the 3D SLAM application. Since the KITTI odometry dataset only provides laser scans. So, we use corresponding sequences from KITTI raw datasets where IMU data is given along with laser scans. For the Ford Campus dataset, two sequences are provided which are both used in this section. In this way, we can conduct a thorough evaluation of our system. First localization

\begin{table*}[hbt!]
	\centering
	\setlength{\tabcolsep}{1.5pc}
	\begin{tabular}{|c|c|c|c|c|c|}
		\hline
		Sequence & Method & RMSE & SD & Mean & Median \\ \hline
		\multirow{2}{*}{00} & Our method & \textbf{7.6581} & \textbf{3.1464} & \textbf{6.9821} & \textbf{6.9396} \\ \cline{2-6} 
		& LOAM & \multicolumn{1}{l|}{13.8885} & \multicolumn{1}{l|}{6.3744} & \multicolumn{1}{l|}{12.3397} & \multicolumn{1}{l|}{10.0319} \\ \hline
		\multirow{2}{*}{01} & Our method & \textbf{21.0776} & \textbf{9.5753} & \textbf{18.7793} & \textbf{19.4772} \\ \cline{2-6} 
		& LOAM & 47.9101 & 29.4392 & 37.8441 & 32.5864 \\ \hline
		\multirow{2}{*}{02} & Our method & \textbf{16.6212} & 9.3645 & \textbf{13.7328} & \textbf{12.2492} \\ \cline{2-6} 
		& LOAM & 19.8842 & \textbf{5.8554} & 19.0034 & 18.2693 \\ \hline
		\multirow{2}{*}{03} & Our method & \textbf{1.6546} & \textbf{0.7379} & \textbf{1.4812} & \textbf{1.2498} \\ \cline{2-6} 
		& LOAM & \multicolumn{1}{l|}{3.6803} & \multicolumn{1}{l|}{2.0770} & \multicolumn{1}{l|}{3.0390} & \multicolumn{1}{l|}{2.1166} \\ \hline
		\multirow{2}{*}{04} & Our method & \textbf{0.9366} & \textbf{0.4709} & \textbf{0.8101} & \textbf{0.7453} \\ \cline{2-6} 
		& LOAM & \multicolumn{1}{l|}{2.7308} & \multicolumn{1}{l|}{1.4745} & \multicolumn{1}{l|}{2.3003} & \multicolumn{1}{l|}{1.7387} \\ \hline
		\multirow{2}{*}{05} & Our method & 4.4767 & 2.5246 & \textbf{3.6972} & \textbf{3.3148} \\ \cline{2-6} 
		& LOAM & \multicolumn{1}{l|}{\textbf{4.3881}} & \multicolumn{1}{l|}{\textbf{1.9872}} & \multicolumn{1}{l|}{3.9126} & \multicolumn{1}{l|}{3.5679} \\ \hline
		\multirow{2}{*}{06} & Our method & \textbf{3.5110} & \textbf{1.3098} & \textbf{3.258} & \textbf{3.0294} \\ \cline{2-6} 
		& LOAM & 3.6839 & 1.9197 & 3.5504 & 3.1453 \\ \hline
		\multirow{2}{*}{07} & Our method & 3.5085 & 1.6567 & 3.0931 & 2.5612 \\ \cline{2-6} 
		& LOAM & \multicolumn{1}{l|}{\textbf{1.8219}} & \multicolumn{1}{l|}{\textbf{0.6903}} & \multicolumn{1}{l|}{\textbf{1.6862}} & \multicolumn{1}{l|}{\textbf{1.5923}} \\ \hline
		\multirow{2}{*}{08} & Our method & \textbf{11.6740} & \textbf{2.1580} & \textbf{11.4728} & \textbf{11.5218} \\ \cline{2-6} 
		& LOAM & 15.0168 & 6.8326 & 13.3840 & 15.9555 \\ \hline
		\multirow{2}{*}{09} & Our method & \textbf{6.3078} & \textbf{2.7865} & \textbf{5.6594} & \textbf{6.5253} \\ \cline{2-6} 
		& LOAM & 7.9374 & 3.0628 & 7.5692 & 7.3235 \\ \hline
		\multirow{2}{*}{10} & Our method & \textbf{5.2820} & \textbf{2.9543} & \textbf{4.3794} & \textbf{3.7817} \\ \cline{2-6} 
		& LOAM & 7.1821 & 3.6101 & 6.2109 & 6.3406 \\ \hline
	\end{tabular}
	\caption{RMSE (root mean square error), SD (standard deviation), Maximum error, Median and Mean of the absolute trajectory error are used to evaluate the performance of our system against LOAM.}
	\label{tab:loam}
\end{table*}
\noindent
accuracy is compared with LOAM, then we do a comparison of accuracy and loop closure performance with the cartographer. Lastly, we evaluate the computational cost of our method, cartographer, and LOAM.

In our implementation, we run the LIDAR data in real-time and feed it to our algorithm pipeline. For the camera intrinsic parameters, we set the image canter at $(0, 0)$ and focal length equal to 500. Extrinsic parameters are specified as the rotation matrix equal to 3x3 identity matrix and translation equal to $[-25, -25, 70]$. The image size is set to 750x750. 

\begin{figure*}[!ht]
	\centering
	\begin{subfigure}{.3\textwidth}
		\centering
		\includegraphics[scale=0.25]{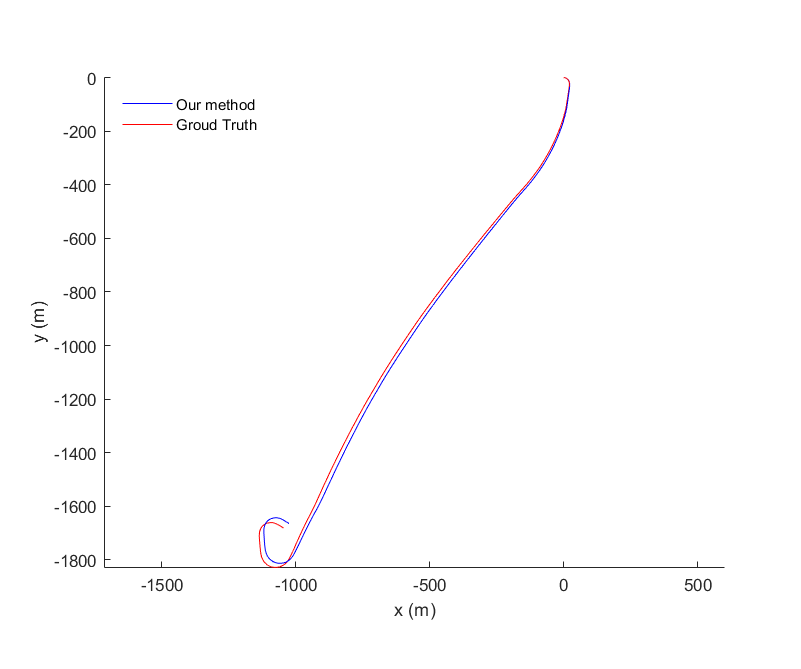}
		\caption{Sequence 01}
	\end{subfigure}\qquad
	\begin{subfigure}{.3\textwidth}
		\centering
		\includegraphics[scale=0.25]{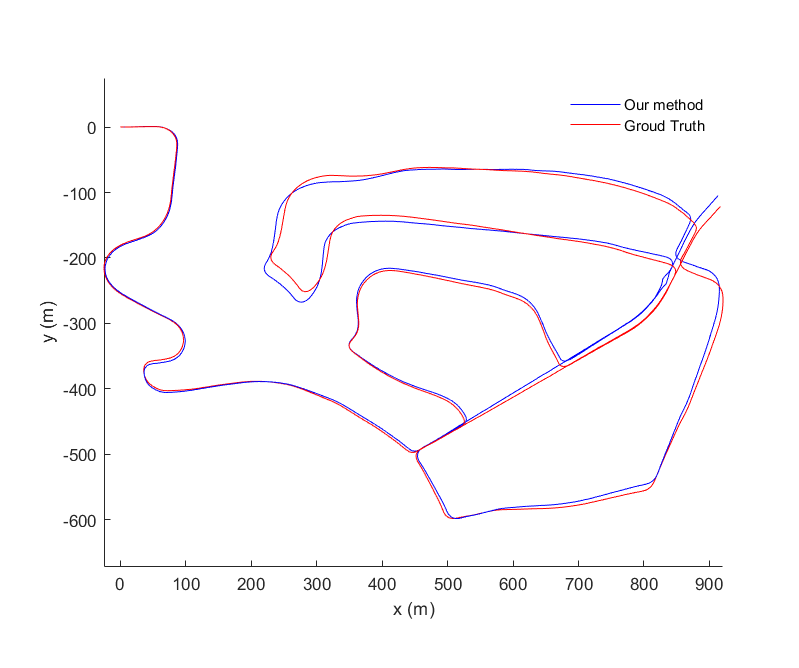}
		\caption{Sequence 02}
	\end{subfigure}\qquad
	\begin{subfigure}{.3\textwidth}
		\centering
		\includegraphics[scale=0.25]{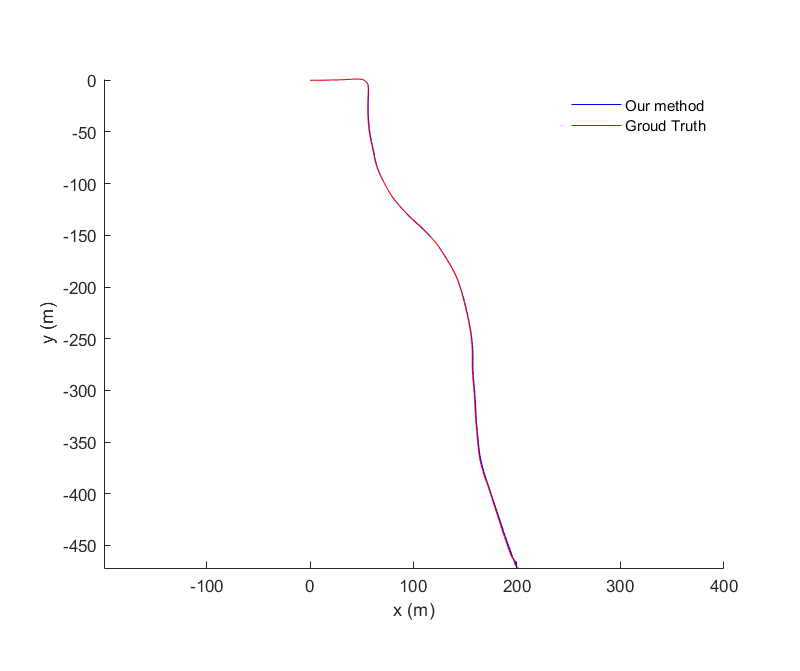}
		\caption{Sequence 03}
	\end{subfigure}\qquad
	\begin{subfigure}{.3\textwidth}
		\centering
		\includegraphics[scale=0.25]{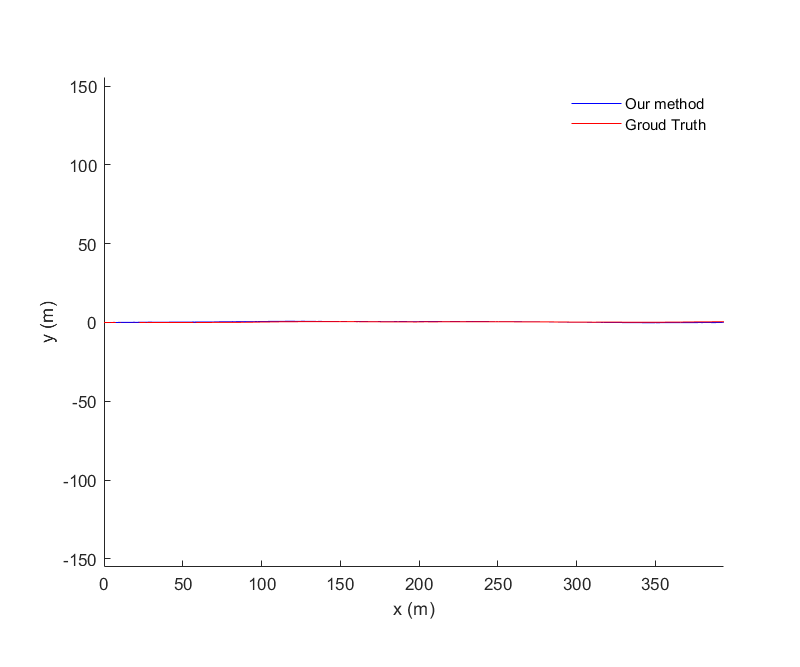}
		\caption{Sequence 04}
	\end{subfigure}\qquad
	\begin{subfigure}{.3\textwidth}
		\centering
		\includegraphics[scale=0.25]{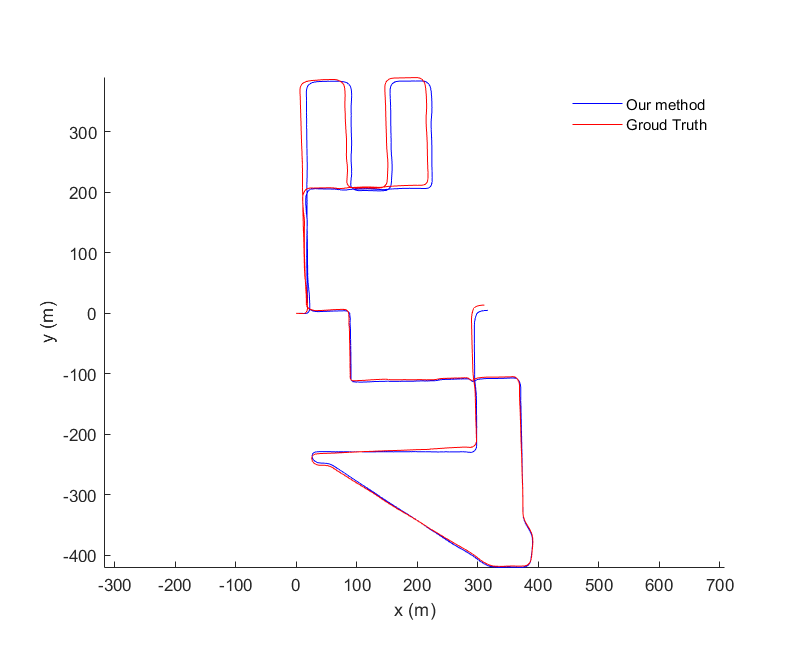}
		\caption{Sequence 08}
	\end{subfigure}\qquad
	\begin{subfigure}{.3\textwidth}
		\centering
		\includegraphics[scale=0.25]{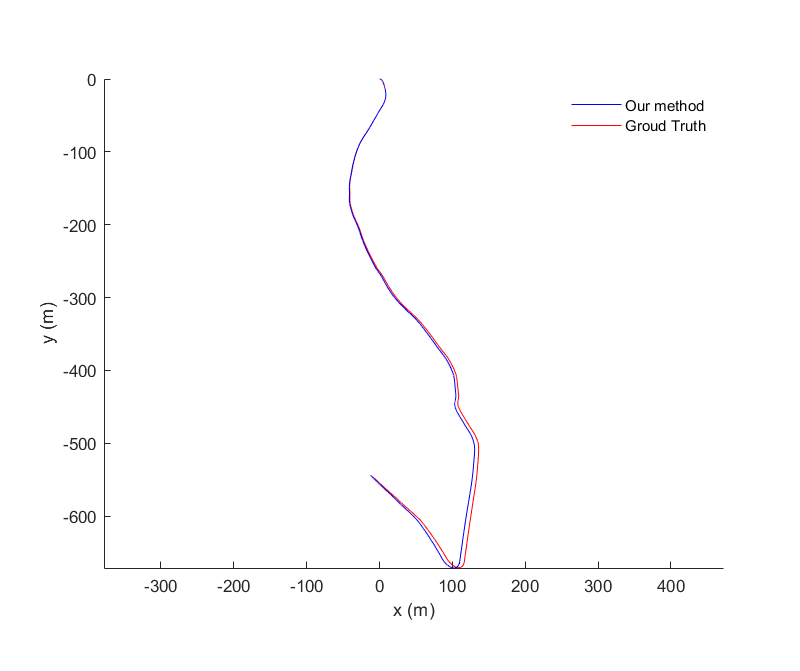}
		\caption{Sequence 10}
	\end{subfigure}
	\caption{Trajectory plots of KITTI dataset training sequences without loop closure}
	\label{fig:kitti_nolc}
\end{figure*} 

\begin{table*}[hbt!]
	\centering
	\setlength{\tabcolsep}{1.3pc}
	\begin{tabular}{|l|l|l|l|l|l|}
		\hline
		Seqeunce & Method & RMSE & SD & Mean & Median \\ \hline
		\multirow{2}{*}{2011\_10\_03\_drive\_0027} & Our method & \textbf{7.6581} & \textbf{3.1464} & \textbf{6.9821} & \textbf{6.9396} \\ \cline{2-6} 
		& Cartographer & 11.6690 & 4.4771 & 10.7761 & 10.5435 \\ \hline
		\multirow{2}{*}{2011\_09\_30\_drive\_0018} & Our method & \textbf{4.4767} & \textbf{2.5246} & \textbf{3.6972} & \textbf{3.3148} \\ \cline{2-6} 
		& Cartographer & 15.2227 & 6.0145 & 13.9849 & 14.3863 \\ \hline
		\multirow{2}{*}{2011\_09\_30\_drive\_0020} & Our method & \textbf{3.5110} & \textbf{1.3098} & \textbf{3.258} & \textbf{3.0294} \\ \cline{2-6} 
		& Cartographer & 25.3995 & 5.103 & 24.8821 & 23.0613 \\ \hline
		\multirow{2}{*}{2011\_09\_30\_drive\_0027} & Our method & \textbf{3.5085} & \textbf{1.6567} & \textbf{3.0931} & \textbf{2.5612} \\ \cline{2-6} 
		& Cartographer & 5.0272 & 3.3951 & 3.7107 & 2.7012 \\ \hline
		\multirow{2}{*}{2011\_09\_30\_drive\_0033} & Our method & \textbf{6.3078} & \textbf{2.7865} & \textbf{5.6594} & \textbf{6.5253} \\ \cline{2-6} 
		& Cartographer & 13.1488 & 6.5532 & 11.4007 & 9.9995 \\ \hline
	\end{tabular}
	\caption{RMSE (root mean square error), STD (standard deviation), Maximum error, Median and Mean of the absolute trajectory error are used to evaluate the performance of our system against cartographer.}
	\label{tab:carto}
\end{table*}

\subsection{Tracking Performance}
To assess the accuracy of vehicle localization by our system we compare the results with state of the art methods LOAM and cartographer. Absolute trajectory error \cite{Ref34} is used to evaluate the performance of complete SLAM system along with loop closure. First, we align the trajectory computed by our system with the ground truth, then estimate absolute error and analyze the performance through mean, median, standard deviation, and root mean square error (RMSE) of the absolute trajectory error. There is an evaluation tool also proposed in \cite{Ref32}, which is used for KITTI odometry benchmark. Relative pose error compares the estimated pose with ground truth for segments of length. 

\begin{figure*}[h]
	\centering
	\begin{subfigure}{.3\textwidth}
		\centering
		\includegraphics[scale=0.25]{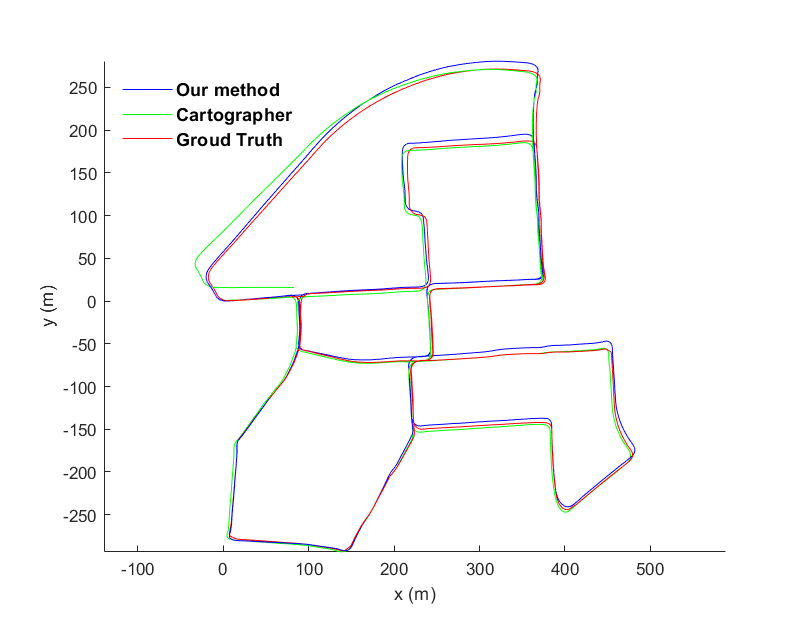}
		\caption{2011\_10\_03\_drive\_0027}
	\end{subfigure}\qquad
	\begin{subfigure}{.3\textwidth}
		\centering
		\includegraphics[scale=0.25]{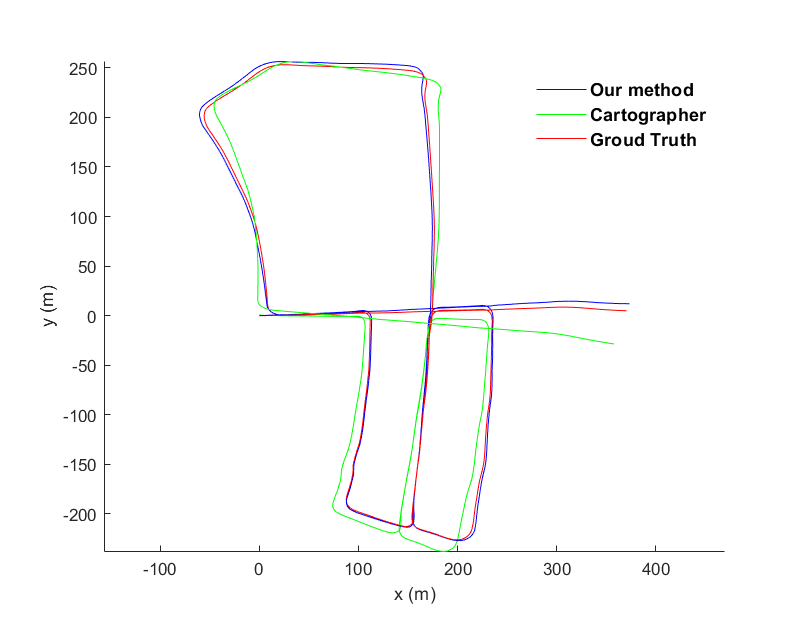}
		\caption{2011\_09\_30\_drive\_0018}
	\end{subfigure}\qquad	
	\begin{subfigure}{.3\textwidth}
		\centering
		\includegraphics[scale=0.25]{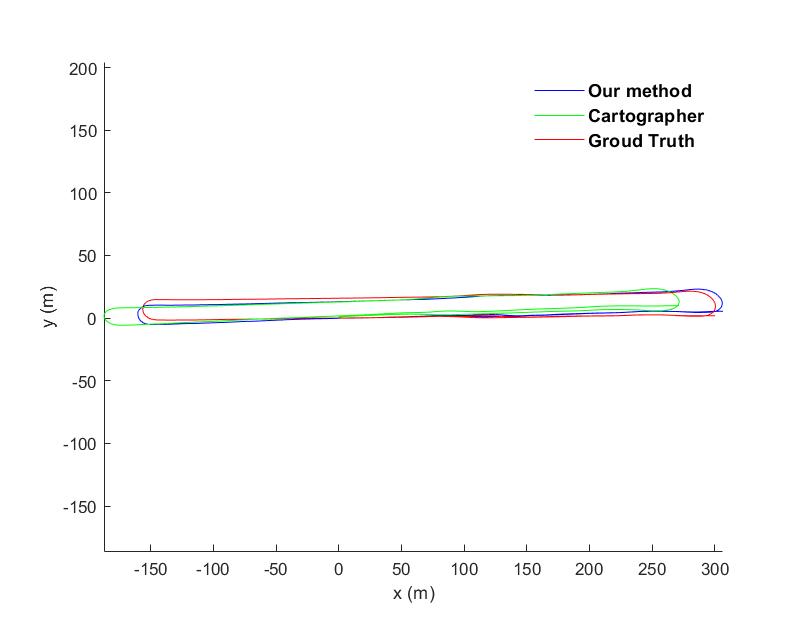}
		\caption{2011\_09\_30\_drive\_0020}
	\end{subfigure}\qquad
	\begin{subfigure}{.3\textwidth}
		\centering
		\includegraphics[scale=0.25]{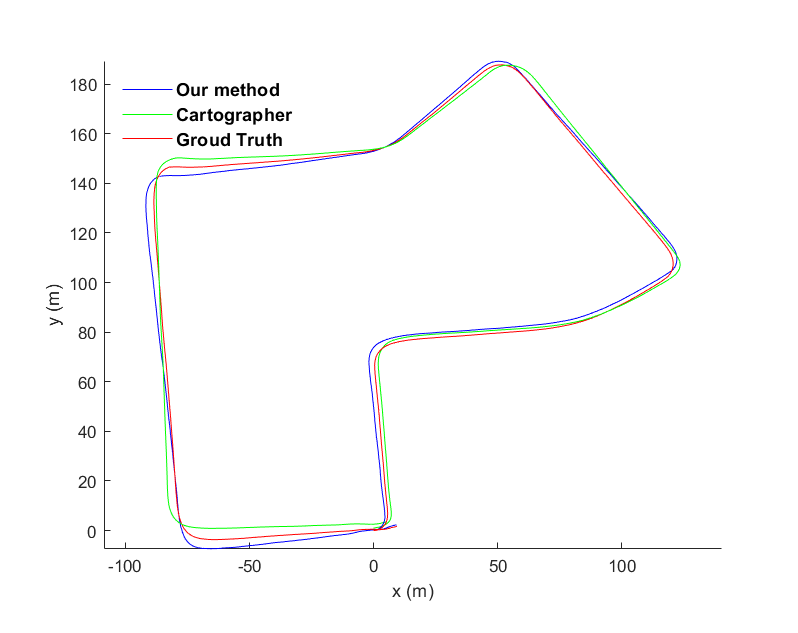}
		\caption{2011\_09\_30\_drive\_0027}
	\end{subfigure}\qquad
	\begin{subfigure}{.3\textwidth}
		\centering
		\includegraphics[scale=0.25]{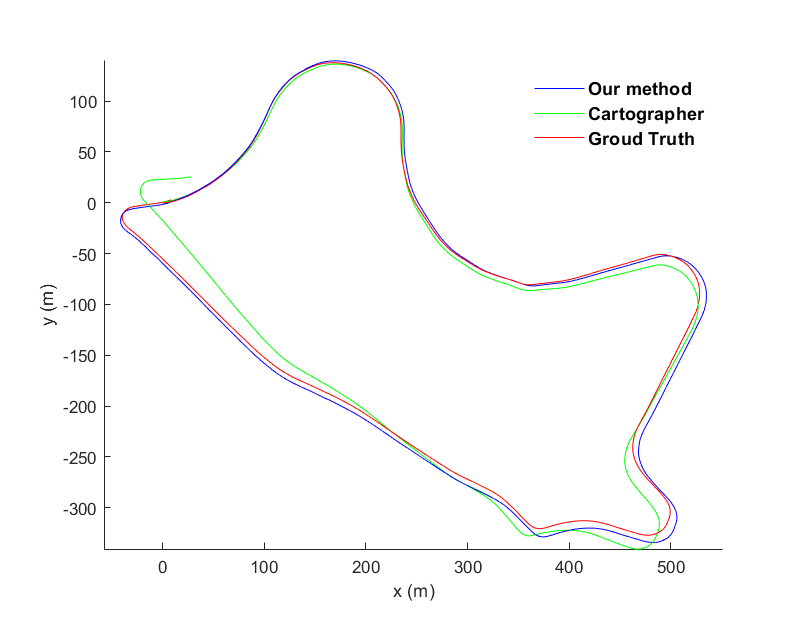}
		\caption{2011\_09\_30\_drive\_0033}
	\end{subfigure}
	\caption{Trajectory plots after loop closure detection and optimization for KITTI sequences.}
	\label{fig:kitti_traj}
\end{figure*}

First, we compare the performance of our system with LOAM. LOAM is also a feature-based method that extracts corner and plane features from a 3D point cloud. We run the open-source implementation of LOAM on the 11 training sequences of the KITTI dataset. Table~\ref{tab:loam} shows the values of all the 11 sequences in terms of mean, median, standard deviation, and root mean square values for the trajectories generated by our method and LOAM. Our system's overall performance is better as compared to LOAM for the training dataset. Figure \ref{fig:kitti_nolc} shows the trajectories for sequences 01, 02, 03, 04, 08 and 10, of our method with ground truth. The trajectories of the remaining 5 sequences along with google cartographer are shown in figure \ref{fig:kitti_traj}. Our method's performance is much better especially in longer sequences such as 00 and 02. The trajectory of LOAM drifts away from ground truth for longer sequences but our method gives better accuracy as a result of loop closure optimization. While for the remaining sequences our accuracy is also better because of multiple levels of optimization employed in this paper. 

Next, we run cartographer on five sequences with loop closure. It requires IMU and lidar data for laser SLAM, which is provided in KITTI raw dataset. Table \ref{tab:carto} shows the ATE values for the five sequences. Our method performs better for all the sequences due to superior loop closure performance. 

We submitted the results of the 11 test sequences to the KITTI benchmark, the translation error for our method is 1.47\% and rotation error is 0.0033 deg/m. The translation error is lower for the test sequences and one of the reasons is the presence of dynamic objects e.g. moving cars on the highway. Such objects, when appearing on the rasterized images may cause outliers in features and affect the performance of the system.

\subsection{Global Bundle Adjustment}

We have implemented global bundle adjustment for loop closure
optimization to give a comparison of the results obtained from the method used in this paper. Table~\ref{gloablBA} shows the results for sequence 00 of the KITTI dataset. We used three parameters to assess the performance of two methods i.e. RMSE error, computational time, and CPU load during optimization. The trajectory error for the global BA is slightly lower than our method. But at the same time, computational time and CPU load are much higher for global BA. In the case of global BA,  we have keyframe poses and map-points as nodes and to optimizes all these nodes increases the computational requirements. For this reason, we preferred using pose graph optimization without map-points. As it gives accurate results at a much lower computational cost.

\begin{table}[h!]
	\centering
	\begin{tabular}{|c|c|c|}
		\hline
		& Global BA & \begin{tabular}[c]{@{}l@{}}Pose graph\\ optimization\end{tabular} \\\hline
		RMSE$\pm$SD & 7.4172$\pm$2.6985m & 7.6581$\pm$3.14m \\\hline
		\begin{tabular}[c]{@{}l@{}}Computational\\ Time\end{tabular} & 15.731s & 0.555s \\\hline
		CPU Load & 26.72$\pm$9.14\% & 13.41$\pm$6.51\% \\\hline
	\end{tabular}
	\caption{A comparison of the results of global BA and pose graph optimization without landmarks for loop closure optimization}
	\label{gloablBA}
\end{table}

\subsection{Loop Closure Performance}

One of the main contributions presented in this paper is a novel loop closure method for laser SLAM i.e. finding nearest key-frames and loop verification based on feature matching. LOAM consists of only tracking and mapping threads, so we use the cartographer to evaluate the efficiency of our loop closure method.  Figure~\ref{fig:kitti_traj} shows the resulting trajectories of our method and cartographer, after pose graph optimization with loop constraints for the KITTI dataset. The two important cases are sequences 2011\_09\_30\_drive\_0027 and 2011\_09\_30\_drive\_0033, in these two sequences, there are only a few frames to detect loop closure at the end of trajectories. In this scenario, the cartographer lagged and was not able to detect loop closure. While we ran our algorithm on these 2 sequences several times and each time our system was able to successfully detect loop closure. 

Sequence 2011\_10\_03\_drive\_0027 is the longest sequence with 4550 frames and contains several loop closures. In this case, when we consider the trajectory of the cartographer, it can detect most of the loops correctly but drifts at the end because of the failure to detect loop closure. The same problem occurs in the result of sequence 2011\_09\_30\_drive\_0018. For our method, we can correct the drift effectively using loop closure. The results show the efficiency of our method in detecting correct loop constraints and minimizing drift in a trajectory through pose graph optimization. Sequence 2011\_09\_30\_drive\_0020 contains long loop closure at the end of the trajectory and it is correctly optimized and corrected by both algorithms. 

There are 2 sequences in the University of Michigan Ford Campus dataset. Only LOAM and our method are tested on the Ford Campus dataset. Figure~\ref{fig:um_traj} shows the result for these 2 sequences. Sequence 1 contains loop closure at the end of the path, the vehicle is driven to its starting position and stopped. Several frames of data are recorded at this position. 

\begin{center}
	\includegraphics[scale=0.35]{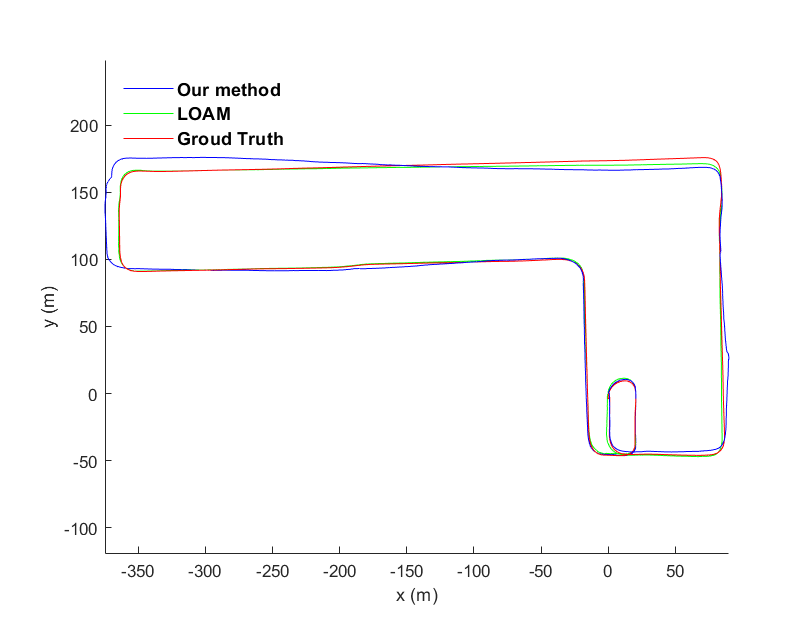}
	\includegraphics[scale=0.35]{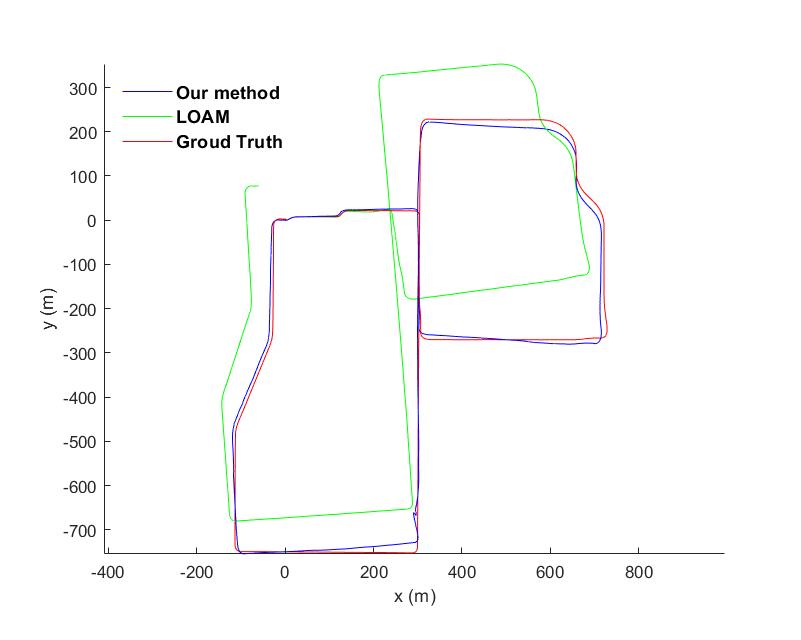}
\captionof{figure}{Trajectory plots for UM Ford Campus dataset sequences.}\label{fig:um_traj}
\end{center}

For sequence-1, there are several frames collected while the car is driving through an empty parking space. In such conditions, it is difficult to extract features, but our method can perform tracking successfully. Loop closure is correctly detected to optimize the trajectory by our method. The trajectory produced by LOAM is closer to ground truth for sequence-1. Sequences-2 is much longer with 2 main loop closure occurrences i.e. one long loop closure in the middle of the path and the second one at the end. Our system is also able to correctly detect a loop in this sequence and correct the drift. For sequence-2, our result is much better as compared to LOAM. Because of the long drive, the trajectory of LOAM drifts away from the ground truth. The loop closure technique presented in this paper works well in different environments proved from the results in our experiments.

\subsection{Computational Efficiency}
One of the main goals of the system presented in this paper is to reduce the computational cost of feature detection from the 3D point cloud and of the whole laser-based SLAM system. To analyze the performance of our system, first, we compare the computational time required for mapping and tracking threads of our method and LOAM. Secondly, a comparison of the time requirements for local SLAM is performed for our method and cartographer. Finally, to give an estimate of the computational load of the complete SLAM algorithm on the system, we also evaluate the CPU load for our method and cartographer.

\begin{table}[h]
	\centering
	\begin{tabular}{|c|c|c|}
		\hline
		Operation & Mean Time (ms) & SD (ms) \\ \hline
		Image Projection & 6 & 3.6 \\ \hline
		\begin{tabular}[c]{@{}c@{}}Feature  Extraction \\ and matching\end{tabular} & 32.5 & 13.1 \\ \hline
		Pose Estimation & 3.2 & 0.98 \\ \hline
		\begin{tabular}[c]{@{}c@{}}Key-frame and\\ map-points registration\end{tabular} & 5.8 & 2.5 \\ \hline
		Local BA & 55.7 & 11.2 \\ \hline
	\end{tabular}
	\caption{Meantime and its standard deviation estimated for Image formation, Tracking and Mapping threads in our system for sequence 00 of KITTI dataset.}
	\label{tab:time}
\end{table}

\begin{figure*}[h!]
	\centering
	\includegraphics[scale=0.3]{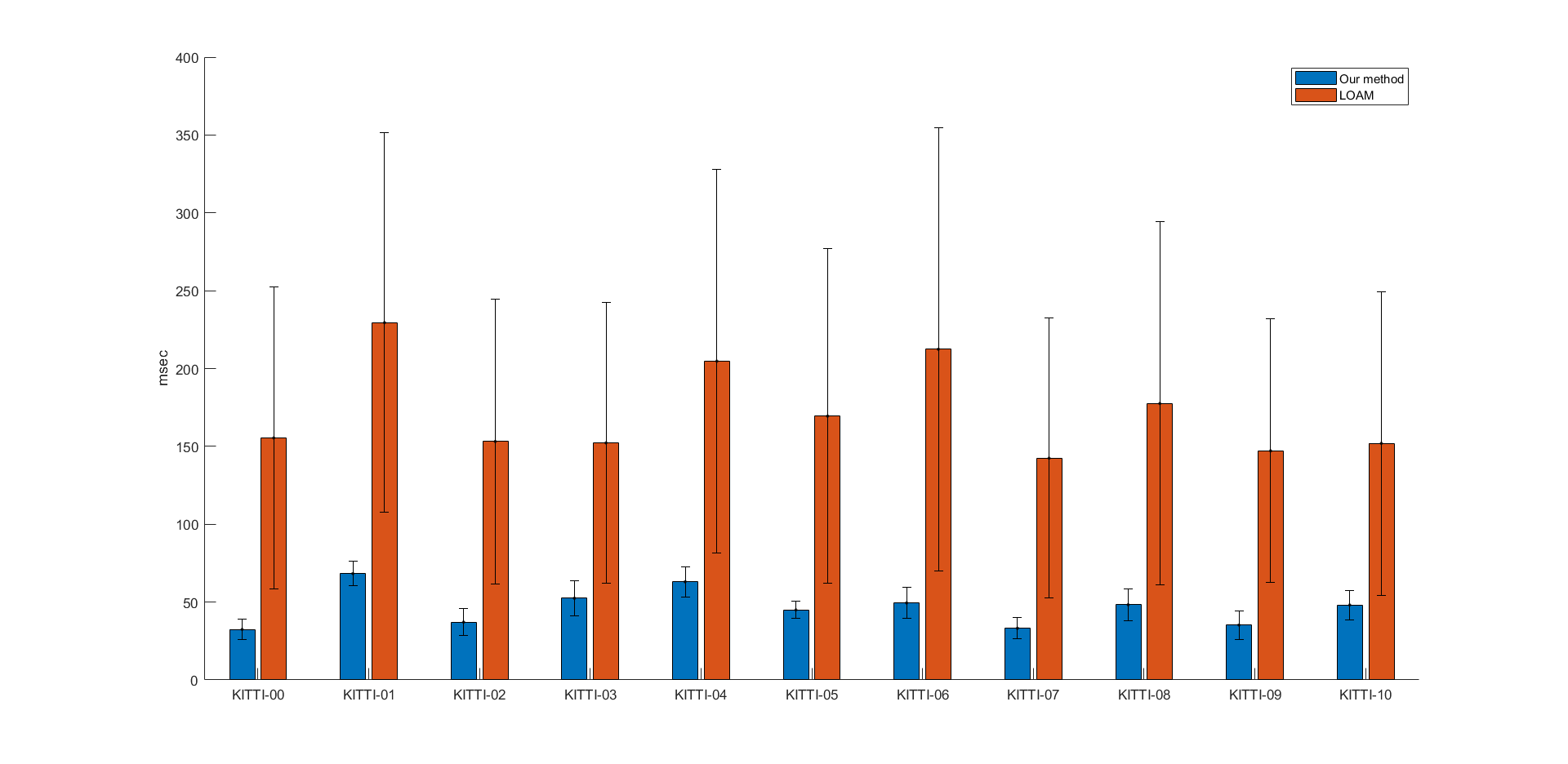}
	\caption{A comparison of time consumed by Tracking thread for our method and LOAM.}
	\label{fig:loam_tracking}
\end{figure*}

\begin{figure*}[h!]
	\centering
	\includegraphics[scale=0.3]{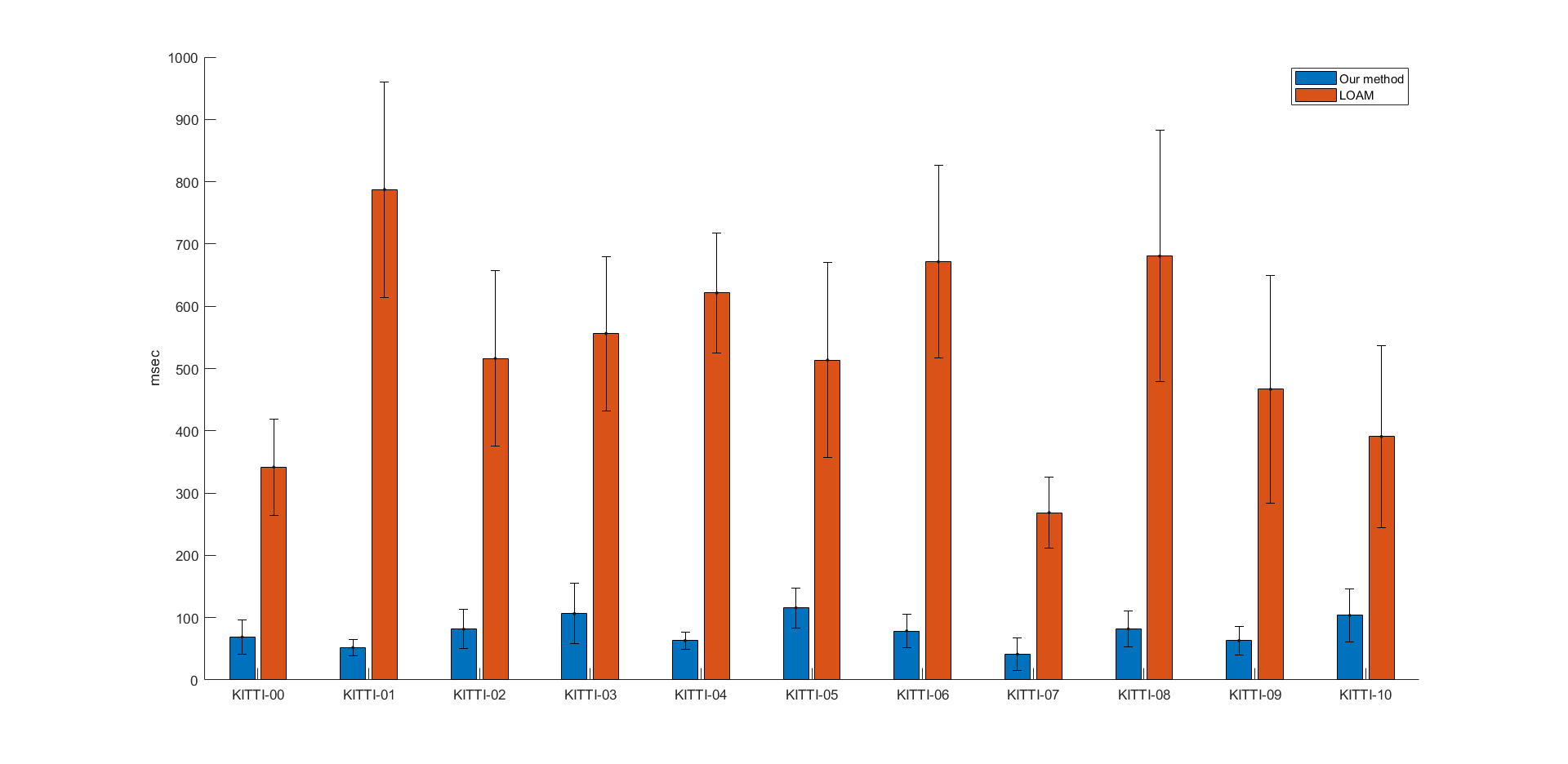}
	\caption{A comparison of time consumed by Mapping thread for our method and LOAM.}
	\label{fig:loam_mapping}
\end{figure*}

Table~\ref{tab:time} presents the values of the time-requirement for each section of our system. For the tracking thread, there are 3 parts i.e. Image projection, Feature extraction and matching, and Pose estimation. The mean time spent on tracking during KITTI dataset experiments is around 41.7 ms. While the mapping thread takes a mean time of 61.5 ms to complete key-frames and map-points registration and performs local bundle adjustment. Figure~\ref{fig:loam_tracking} and figure~\ref{fig:loam_mapping} show a comparison of the meantime and standard deviation for tracking and mapping threads for our system and LOAM. Our system performs these tasks in much lesser time. For Tracking, the time consumption is reduced due to our feature extraction method. LOAM algorithm extracts corner and planner features by scanning through the 3D point cloud. But for our system, we only need to scan a rasterized image at each frame to extract feature points, which substantially reduced the time-requirement to perform tracking. For the Mapping thread, we rely on key-frame structure and apply windowed optimization locally to make our system computationally efficient.

Next, we evaluate the computational efficiency of our method with comparison to the cartographer. The structure of our algorithm is similar to the cartographer in one aspect that it is divided into local and global SLAM threads.  Local SLAM is responsible for tracking the vehicle and builds sub-maps for the cartographer and our algorithm, it includes tracking the vehicle, manages local maps and performs local BA. So, we compare the computational costs of our method and cartographer in two ways i.e. estimate and compare the computational time required for local SLAM and a comparison of CPU load for running complete algorithms. In this way, we can get a comprehensive assessment of the computational cost for both methods.

\begin{figure}[htp!]
	\centering
	\includegraphics[scale=0.3]{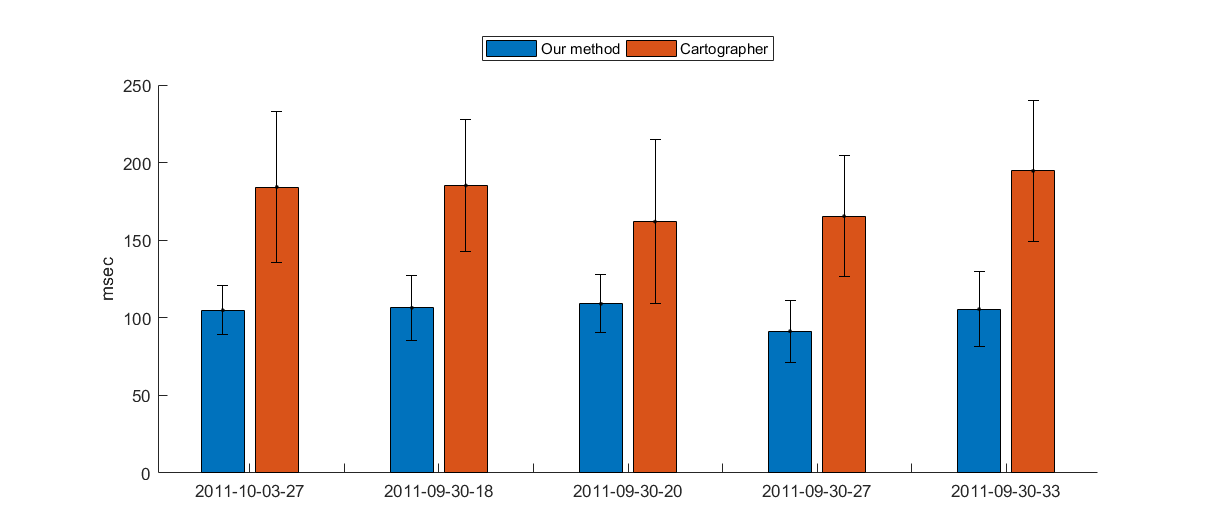}
	\caption{A comparison of the time consumed by local SLAM for our method and cartographer.}
	\label{fig:carto_time}
\end{figure}

First, we look at the time requirements of both methods for local SLAM during experiments on the KITTI dataset. Figure~\ref{fig:carto_time} shows the mean time required for local SLAM calculations along with standard deviation. Our method takes much lesser time across all experiments. We use features for pose estimation while the cartographer relies on direct scan matching. Secondly, we use a key-frame structure in local mapping which makes our system more efficient.

Second, a comparison of the CPU load for both systems is given in Table~\ref{tab:carto_cpu}. The values from the table show resources required by each system to run a full SLAM pipeline including local SLAM and global optimization based on loop closure search. It is clear from the results that the cartographer requires more resources to run the complete SLAM algorithm.  

\begin{table}[h!]
	\centering
	\begin{tabular}{|l|l|l|}
		\hline
		& Cartographer & Our method \\ \hline
		10\_03\_27 & 32.62$\pm$9.15 & \textbf{13.41$\pm$6.51} \\ \hline
		09\_30\_18 & 33.6$\pm$8.25 & \textbf{8.51$\pm$3.54} \\ \hline
		09\_30\_20 & 26.55$\pm$11.94 & \textbf{5.35$\pm$1.58} \\ \hline
		09\_30\_27 & 18.59$\pm$4.78 & \textbf{5.16$\pm$1.74} \\ \hline
		09\_30\_33 & 23.03$\pm$5.7674 & \textbf{6.58$\pm$3.12} \\ \hline
	\end{tabular}
	\caption{Mean percentage values of CPU usage by cartographer and our method}
	\label{tab:carto_cpu}
\end{table}

For cartographer, which relies on scan matching for pose estimation and then search through sub-maps for loop closure detection making its computational load to rise. Our system relies on a simple and cost-effective approach to solve the SLAM problem. First, the computational requirements for local SLAM are cut by using features, which are easier to extract using rasterized images of 3D point cloud and secondly key-frames structure keeps local BA efficient. For global SLAM which is responsible for loop closure detection and pose-graph optimization, it is based on a simple approach to search for loop closure. These features ensure that CPU load and time consumption for each step of our algorithm remains minimum.

\section{Conclusion}

Feature detection from a 3D point cloud presents many advantages such as lesser computational cost and more accurate state estimation. But it can be a challenging task. In this paper, we presented a feature detection approach by first projecting a 3D point cloud on a virtual plane to form an image and then extract ORB features from these images. We then estimate the 6DOF pose and register map using these features. In our system, the trajectory and map-points are first optimized using local bundle adjustment and then using pose graph optimization with loop closure constraints. The performance of our system is validated through implementation on KITTI and the University of Michigan Ford Campus datasets. We showed that our system can produce an accurate result with a substantial reduction in computational cost. One issue seen in experiments is the presence of dynamic objects in the environments can affect the performance of the system. Our next research goal is to design an efficient mapping strategy, that can account for both static and dynamic objects to further improve the accuracy of the system.

\bibliography{references}

\end{document}